# Knowledge Representation for Lexical Semantics: Is Standard First Order Logic Enough?

Marc Light and Lenhart Schubert

August 22, 1994

## 1 Introduction

Natural language understanding applications such as interactive planning [1] and face-to-face translation [20] require extensive inferencing. Many of these inferences are based on the meaning of particular open class[1] words. Providing a representation that can support such lexically-based inferences is a primary concern of lexical semantics.[2] The representation language of first order logic (FOL) has well-understood semantics and a multitude of inferencing systems have been implemented for it. Thus it is a prime candidate to serve as a lexical semantics representation. However, we argue that FOL, although a good starting point, needs to be extended before it can efficiently and concisely support all the lexically-based inferences needed.

Most lexical semantics representation systems utilize either KL-ONE-inspired terminological logics [6, 2, 15, 21] or typed feature structure (TFS) logics [11, 10]. Representationally, terminological logics are subsets of FOL [29, 35, 7, 36, 14] as are TFS logics [29, 19, 18, 37].[3] Thus, we suggest that lexical semanticists interested in supporting lexically-based inferences need to look for ways to enrich their representational systems. We are not alone in this suggestion (see [8, 32, 26]). However, to our knowledge the specific extensions we suggest are novel to the lexical semantics literature.

Most of the examples on which we base our arguments are from an interactive planning dialogue understanding project: the TRAINS project [1]. The goal of the TRAINS project is to build a system that can assist a human manager who is attempting to solve a planning problem. The domain includes trains, rail connections, goods, cities, factories, etc. The computer system will have knowledge about the current state of the world, schedules, timetables, and other relevant information and will interact with the manager in spoken English. A typical planning problem would be to deliver 1000 gallons of orange juice to a specific city by a certain time. To solve this problem the manager would be assisted by the system in scheduling the delivery of the oranges to the orange juice factory and the subsequent shipping of juice to the designated city.

---

[1]Open class words are those that are either adjectives, adverbs, nouns, or verbs. Closed class words are prepositions, determiners, conjunctions, etc.

[2]Lexical semantics is also concerned with issues besides supporting inference. A prominent line of research in lexical semantics concerns itself with the link between a verb's meaning and its syntactic characteristics (see [23]).

[3]It should be noted that many of the systems that use TFS logics, view their TFS representations as descriptions/short-hand for a more expressive semantic representation not as the representation itself [31, 10, 33]. The argument presented here is compatible with this position.

A prototype of the system has been implemented (see [1]). In addition, actual dialogues have been collected in which the role of the system is played by a human in order to determine what natural dialogues are like (see [13]). By relying mostly on examples taken from these dialogues, we illustrate the relevance of the issues we address to mundane, naturally occurring discourse. Moreover, the dialogues provide a task-oriented context in which it is generally clear what inferences are required for understanding a given utterance; thus they provide a more constrained framework for semantic theorizing and experimentation than unrestricted texts or dialogues.

Before we can proceed we need to specify what are and what are not lexically-based inferences. A lexically-based inference is one that depends on a lexical axiom. A lexical axiom is one that involves a semantic atom that is the translation of an open class word (assuming a meaning postulate approach). The following axiom is a lexical axiom and links the verb *enter* with its result state.

(1) $\forall x \forall y [\texttt{result-state}(\texttt{enter}(x,y)) \supset \texttt{contained-in}(x,y)]$

Using it, we could make a lexically-based inference from *the boxcar entered the factory* that the boxcar is in the factory. Furthermore, this inference is based on the word *enter* and not on the word *boxcar*. Such inferences can be contrasted with "structural" inferences such as in (2):

(2) there are at least three cities with orange juice factories and large train stations $\rightarrow$
there are at least three cities with orange juice factories

This depends on properties of certain classes of logical operators, specifically the class of upward monotone quantifiers [4] and the conjunction operator, rather than on the lexical semantics of specific open-class words. Note that if one substitutes the downward monotone quantifier *fewer than three* for *at least three* the inference no longer follows.

Besides distinguishing lexically-based inferences from structural ones, we also intend to distinguish them from inferences based on world knowledge. Without getting too embroiled in the issue of whether there is a formal and sharp distinction between knowledge about lexical meanings and world knowledge, we want to identify lexical knowledge with the sort of knowledge usually treated as terminological knowledge in KR systems. Ultimately a formal distinction may require use of a necessity-like modal operator in axioms like (1), to capture the truth of such axioms in all worlds (or situations), but we set aside this issue here.

## 2 Extensions to FOL needed to support lexically-based inferences

In this section we will introduce a number of extensions to FOL and provide examples that motivate them. More specifically, one should add restricted quantification and non-standard quantifiers, modal operators, predicate modification, and predicate nominalization. These representational tools are available in some systems for sentence-level semantics [16, 28, 3].

It should be noted at the outset that each example used to motivate an extension can be handled in FOL. However, the use of FOL leads to complex and unnatural paraphrases of intuitively simple facts, makes the encoded knowledge harder for system developers to comprehend and modify, and complicates inference. By adding a small amount of expressive power, concise and comprehensible representations can be given which facilitate efficient inferencing.

In our examples, we distinguish *direct* and *indirect* motivation of specific extensions to FOL; i.e., a lexical item may directly correspond to a type of operator (such as predicate modifiers) unavailable in FOL, and it may indirectly involve nonstandard operators through its axiomatization. Our argumentation is necessarily sharply abridged for this abstract.

### 2.1 Non-standard quantifiers and restricted quantification

We begin with an extension for which we cannot yet muster much evidence from TRAINS, but for which there are good reasons from a more general perspective and which already enjoys rather broad acceptance.

Previously, when circumscribing lexically-based inferences, we mentioned upward monotone quantifiers. These include *at least three* (see (2)), *all, a few, most*, etc. Such examples motivate the augmentation of FOL with corresponding nonstandard quantifiers; and the nominals with which they combine (as in *at least three cities with orange juice factories*) motivate the inclusion of formulas restricting the domains of the quantified variables.[4]. By utilizing these extensions, the following axiom enables inferences like the one in (2) to be made efficiently.

---

[4]A number of terminological logics are sorted logics (*e.g.* [32]). In sorted logics, the domain of a variable is restricted by the variable's sort. Representationally, this is much like restricted quantification, though more limited.

(3) For all upward monotone quantifiers $Q$ and all predicates $P_1$, $P_2$, and $P_3$:
    $Q\ x : P_1(x)\ [P_2(x) \land P_3(x)] \supset Q\ x : P_1(x)\ [P_2(x)]$

Note that no reasoning about cardinality is required.

As a direct motivation for a nonstandard quantifier syntax, the above argument pertains only to the closed category of determiners (in combination with certain adverbs and numeral adjectives). However, it it clear that this syntax will also simplify the axiomatization of many open-class words in future extensions of the TRAINS vocabulary. As a constructed (but uncontrived) example consider (4) and assume that the system has in its knowledge base that the majority of cars are tankers and are in Elmira

(4) M: Are the *majority* of the cars tankers?

It should be able to infer that the majority are tankers. By using the axiom below and a suitable treatment of conjunction, the system could make use of the axioms for upward monotone quantifiers.

(5) $\forall a, b\ \mathtt{majority}(a, b) \supset \mathtt{Most}\ z : [z \in a]\ [z \in b]$

In the above axiom we assume that $a$ and $b$ denote collections and that $\in$ has been appropriately axiomatized.

Some other words that would benefit from non-standard quantifiers and restricted quantification are *scarce, rare, minority, scant,* and *predominate*. We also expect that degree adjectives such as *expensive, difficult,* or *intelligent* will require axiomatizations involving nonstandard, restricted quantifiers. For example, a *difficult problem* in the TRAINS domain is one that exacts more time and effort from the problem solver(s) than *most problems in this domain*. Similarly *dispositional* adjectives such as *perishable* or *fragile* and frequency adverbs such as *usually* also call for restricted nonstandard quantification in their axiomatization, but we omit further details here.

## 2.2 Modal operators (or modal predicates)

Standard FOL has difficulty representing necessity, possibility and propositional attitudes.

Yet examples like (6) and (7) involve adverbs that are most naturally viewed as modal operators:

(6) M: That will probably work

(7) M: Maybe we'll get lucky again

In its context of occurrence, the second sentence refers to the possibility that all the orange juice needed for certain deliveries already exists, obviating the need for orange juice production. It would clearly be hazardous for the system to ignore the adverbs, turning mere wishful thinking into fact!

Such examples provide direct motivation for allowing modal operators in lexical semantics. The argument is weakened by the fact that modal adverbs are somewhat marginal as an open class of lexical items; but we can also argue from adjectives such as *reasonable, reliable, correct* and *right*, and verbs such as *found out that ..., said that ..., would like to ..., make sure ..., trying to ..., wonder if ..., believe*, and *assume*. We restrict our further comments here to some observed uses of *correct* and *reliable*. For instance, in the following request for confirmation, the system should interpret *correct* as applying to the proposition that the time is 2 pm:

(8) M: The time is two pm – is that correct?

Now if the system believes that the time is indeed 2 pm, it should surely infer an affirmative answer to the question – i.e., that it is *correct* that the time is 2 pm. Thus for the relevant sense of *correct*, the lexical semantics should tell the system that for any closed formula $\phi$,

(9) $\texttt{correct}(\phi) \equiv \phi$.

If we adopt such a schema for the meaning of *correct*, we are treating it as a modal operator. An alternative is to assume that *correct* is a predicate, but one that applies to *propositions*. In turn, such an approach calls for the introduction of a reifying operator (such as *that*) for converting sentence contents (propositions) into individuals, allowing their use as predicate arguments. In either case, we are introducing a modal extension to FOL. The case of *reliable* is similar but more subtle. In actually occurring examples this property is often ascribed to items of information:

(10) M: That's reliable information

Intuitively, reliable information is not necessarily correct, though it is necessarily well-founded (i.e., there are good reasons for the presumption of truth). So the axiomatization is less trivial than (9) (and we omit details here), but it still calls for use of a modal operator or modal predicate in the same way.

Concerning indirect motivation for modals, an interesting example is *compatible (with)*, as used in

(11) M: So that sounds like a good temporary plan – let's see if it's compatible with our next objective here which is to deliver a boxcar of bananas to Corning

In order for the plan in (11) to be compatible with the additional banana delivery, it must be *possible* to realize both action types (within the given temporal and other constraints). In general,

(12) $\forall x, y[[\texttt{action-type}(x) \wedge \texttt{action-type}(y) \wedge \texttt{compatible-with}(x,y)] \supset$
$\Diamond \exists x', y'[\texttt{realize}(x', x) \wedge \texttt{realize}(y', y)]]$ .

(We comment on action types in a later subsection.) The semantics of the modal operator $\Diamond$ requires a model structure with either possible worlds (*e.g.* [27]) or situations (*e.g.* [5]). $\Diamond \Phi$ entails that there exists a situation or possible world sufficiently connected to the current one where $\Phi$ is true. (Cf. Dowty's use of the $\Diamond$ operator in his treatment of the semantics of the suffix *-able* [12].)

## 2.3 Predicate modification

By predicate modification we mean the transformation of one predicate into another. Within a general setting for language understanding, we could most easily make the case for allowing predicate modifying operators by pointing to nonintersective attributive adjectives such as *former, cancelled, fake, supposed, simulated*, or *ficticious*. For instance, applying *cancelled* to an event nominal such as *trip* yields a new predicate which is *not* true of actual trips, and so should not be analysed as a conjunction of *cancelled* and *trip*. However, such adjectives do not occur in the TRAINS dialogues collected so far, and we will instead use certain verbs (*make, get, look, sound, seem, begin, construct*) as direct motivation for predicate modifiers.

For instance, the dialogues contain instances where the manager asks

(13) M: Does that sound reasonable?

(referring to a plan), or comments

(14) M: Problem number two looks difficult.

Now a plan can *sound* reasonable even if more careful analysis reveals it to be *un*reasonable. So the system should realize that an affirmative response to the query merely requires the absence of *obvious* flaws in the plan (detectable with limited inferential effort), rather than an actual proof of correctness.

One could attempt to handle such locutions by decomposing them into more complex modal patterns; e.g., $x$ sounds $P$, for $P$ a predicate, might be decomposed into something like "When one considers $x$, one (initially) feels that $x$ is $P$". This is precisely the strategy that has often been suggested for for intensional verbs such as *seeks*. But while plausible definitions (decompositions) exist for some intensional verbs, they are very difficult to contrive for ones like *resemble* (as in *The one-horned goat resembled a unicorn*) or *imagine*. A more general, straightforward approach is to add predicate modifiers to FOL. Thus the translation of *sounds* (when it takes an adjectival complement) would be a predicate modifier, whose meaning is *constrained* – but not *defined* – by axioms like the following:[5]

(15) For all monadic predicates $P$:
$\forall x \, \texttt{sounds}(P)(x) \equiv$
$\forall s, t[\texttt{person}(s) \wedge \texttt{consider}(s,x,t) \supset \texttt{feel-that}(s, P(x), \texttt{end-of}(t))]$

where we are neglecting various subtleties for the sake of brevity.[6] Thus to answer (13), the system would make use of (15) to infer that it need only "consider" the plan in question, until it "feels-that" (i.e., *tentatively* concludes that) the plan is reasonable or otherwise.

Finally, we mention a third class of examples directly motivating predicate modifiers, namely certain VP adverbs such as *almost, nearly* and *apparently*. Again, these do not appear (as yet) in our corpus, but of course are common in other corpora. For example, the "pear stories" of Chafe [9] contain examples such as "[the boy on the bicycle] *almost* ran into a girl", where the desired inference is that he did *not* run into her, but came very close to her.

## 2.4 Predicate nominalization

By predicate nominalization we mean the formation of terms (denoting individuals in the domain of discourse) from predicates (denoting classes of objects, actions or events). In other words, predicate nominalization involves the *reification* of properties (inluding *kinds/species*, *action types*, and *event types*) by application of nominalizing (reifying) operators.

Our line of argument for allowing such operators in the logic employed for lexical semantics is less direct (but we hope no less convincing) than for the previous extensions. We claim that (1) many lexical entries correspond to predicates with one or more arguments ranging over kinds of things, properties, and actions/events (this already came up incidentally in (11)); and (2) the lexical axioms describing these entries will either explicitly involve nominalized predicates or require the substitution of nominalized predicates when used for inference.

---

[5] Note that an equivalence need not be definitional. For instance, a triangle is a polygon whose interior angles add up to 180 degrees – but that is not its definition.

[6] This is, of course, the Montagovian approach, though we are dispensing with Montague's intension operator (writing $\texttt{sounds}(P)(x)$ rather than $\texttt{sounds}(^\wedge P)(x)$) by relying on a slight departure from standard intensional semantics that treats the world (or situation) argument as the last, rather than first, argument of the semantic value of a predicate [17].

As examples of a variety of lexical predicates over (reified) action types, consider the italicized words in the following TRAINS excerpts. We have underlined corresponding action-type arguments where these are explicitly present.

(16) M: What is the best *way* for me to accomplish my *task* ...
(17) S: That's a little beyond my *abilities*
(18) S: The *way* it's going to work is, engine E2 is going to go to city E ...
(19) S: Our current *plan* is to fill ... tankers T3 and T4 with beer ...
(20) S: One other *suggestion* would be that you take the other tanker which isn't being used ...
(21) M: That's not gonna *work*
(22) S: Well that will *delay* departure
(23) S: Right, we can *begin* production ...
(24) M: ... send it off on a particular route and *do* it several times

Clearly *way, task, plan*, and *suggestion* as used in (16 - 20) are predicates over types of actions or events, as the underlined arguments confirm. For instance, the action descriptions underlined in (18) and (19) do not refer to particular future actions at particular times, but to types of actions whose eventual *realization* is hoped to solve the problem at hand. (And the ability deictically referred to in (17) is the ability to specify the best way for the manager to accomplish the current task in (16) – again an action type.) Similarly (21 - 24) illustrate verbs whose subject or object ranges over action types. Note for instance in (22) that a *particular* departure event cannot be delayed – particular events have fixed times of occurrence, but event types in general do not. Likewise in (24), only an action type, not a particular action, can be done "several times".

Similarly the following excerpts contain predicates over kinds/species, again with corresponding arguments underlined:

(25) M: One boxcar of oranges is *enough* to produce the required amount of orange juice
(26) M: And *fill* two tankers *with* beer
(27) M: There's [an] unlimited *source of* malt and hops ...

Note that in (25) the underlined subject of *enough* refers to a *kind* of load or quantity, not to any *particular* load. Similarly the underlined objects of *fill with* and *source of* in (26) and (27) are *kinds* of stuff, not particular realizations of them. (In fact, no particular batch of malt and hops could be "unlimited".)

Turning to the second step of our argumentation, concerning the explicit occurrence of nominalization operators in argument positions of predicates like those above, one very brief example will have to suffice here. Consider the sense of *do* with an action type as object, as in (24). Now to understand (24), the system will have to substitute a term for the action type, "send [the train] off on a particular route", for the pronoun. To infer any further consequences, it will need a meaning postulate something like the following:

(28) For all monadic action predicates $P$:
    $\forall x\ \mathtt{do}(\mathtt{Ka}(P))(x) \supset P(x)$

where $\mathtt{Ka}$ reifies an action predicate (in this case, "send [the train] off on a particular route"). It can then apply semantic and world knowledge about $P$ to draw conclusions about the effects of $P(x)$ (in this case that the train will follow the route in question and reach its destination).

## 3  Conclusion

A popular goal for the future is to build intelligent agents that we can communicate with using natural language. If this goal is to be attained such agents will have to be able to perform complex inferencing. This will require lexicons that can support extensive lexically-based inferencing. In order to support these inferences, representations for lexical semantics will have to be richer than they are now. We have provided motivations for particular extensions, drawing many of the illustrations from actual dialogues in the TRAINS domain – a "practical" domain of the sort for which we can realistically endeavor to build an intelligent assistant.

Fortunately, the extensions of FOL for which we have argued are not new (as noted at the outset). Indeed, they are a subset of the extensions that are available in Episodic Logic (EL) [16, 17], a logic designed to be expressively and inferentially adequate as both a logical form for natural language and as a general representation for commonsense knowledge. Similar formalisms are those used in the Core Language Engine [2, 3] and Nerbonne and Laubsch's NLL [22, 30]. EL is an intensional, situational extension of FOL that provides a systematic syntax and formal semantics for sentence and predicate nominalization (reification), sentence and predicate modification, nonstandard restricted quantifiers, $\lambda$-abstraction, and pairing of arbitrary sentences with situation- (episode-) denoting terms (where those sentences are interpreted as describing or characterizing those situations).

Inference in EL has been shown to be practical through the EPILOG implementation [34], with examples ranging from fairy tale fragments and aircraft maintenance reports [16, 17] to the Steamroller theorem-proving problem. As well, EL as been used as the front-end logical form in the TRAINS system [1]. A gratifying conclusion from the text understanding experiments is that increased expressiveness often simplifies inference, allowing conclusions to be drawn in one or two steps that would require numerous steps in an FOL "reduction" of the same information. Finally, EL has been used as a representation for forming hypotheses about meanings of derived words, such as *reload* (given a lexical entry for *load*), and *performance* (given a lexical entry for *perform*) [24, 25]. As might be expected, the representational requirements for expressing such hypotheses are very similar to those we have pointed out here.

By allowing the same flexibility in the representation of lexical semantics as is provided by logics like EL for sentence-level semantics, we should be able to combine lexical semantic knowledge with world knowledge in a uniform, integrated fashion to achieve understanding at least in restricted, task-oriented domains.